\documentclass[10pt,twocolumn,letterpaper]{article}

\usepackage{cvpr}              %

\usepackage{xcolor}[dvipsnames]
\usepackage{graphicx}
\usepackage{amsmath}
\usepackage{amssymb}
\usepackage{mathtools}
\usepackage{booktabs}
\usepackage{enumitem}
\usepackage{xspace}

\usepackage[pagebackref,breaklinks,colorlinks]{hyperref}

\usepackage[accsupp]{axessibility}  %

\usepackage[capitalize]{cleveref}
\crefname{section}{Sec.}{Secs.}
\Crefname{section}{Section}{Sections}
\Crefname{table}{Table}{Tables}
\crefname{table}{Tab.}{Tabs.}

\def\cvprPaperID{1193} %
\def\confName{CVPR}
\def\confYear{2023}

\newcommand{\edi}{University of Edinburgh}
\newcommand{\adobe}{Adobe Research}
\newcommand{\name}{RenderDiffusion\xspace}

\newcommand{\cam}[1]{#1}

\newcommand{\fig}[1]{Fig.~\ref{fig:#1}}
\newcommand{\tab}[1]{Tab.~\ref{tab:#1}}

\newcommand*{\affmark}[1][*]{\textsuperscript{#1}}
\newcommand*{\affaddr}[1]{#1} %

\newcommand{\boldstart}[1]{\noindent\textbf{#1}}
\newcommand{\boldstartspace}[1]{\vspace{0.1in}\noindent\textbf{#1}}

\begin{document}

\title{\name: Image Diffusion for 3D Reconstruction, \\ Inpainting and Generation}

\author{Titas Anciukevičius\affmark[1,2],
Zexiang Xu\affmark[2],
Matthew Fisher\affmark[2],\\
Paul Henderson\affmark[3],
Hakan Bilen\affmark[1],
Niloy J. Mitra\affmark[2,4],
Paul Guerrero\affmark[2]\\
\affaddr{\affmark[1]\edi{}}, \affaddr{\affmark[2]\adobe}, \affaddr{\affmark[3]University of Glasgow}, \affaddr{\affmark[4]UCL}\\
}

\twocolumn[{%
\renewcommand\twocolumn[1][]{#1}%
\maketitle
\thispagestyle{empty}
\begin{center}
    \vspace{-22pt}
     \includegraphics[width=\linewidth]{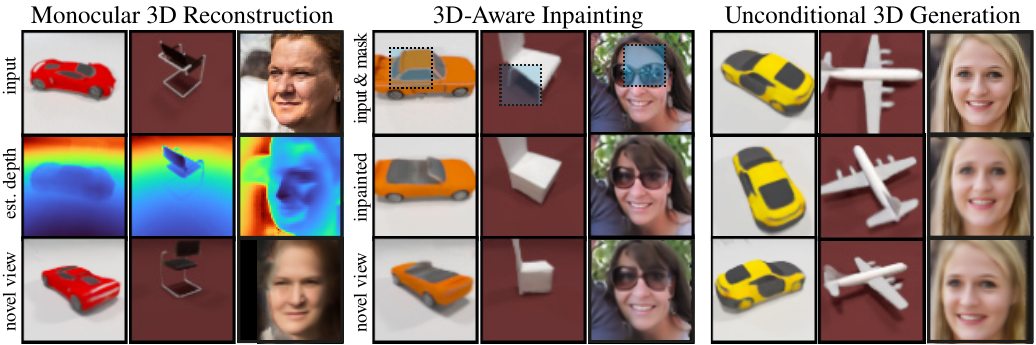}
     \vspace{-17pt}
     \captionof{figure}{We propose a 3D-aware image diffusion model that can be used for monocular 3D reconstruction, 3D-aware inpainting, and unconditional generation, while being trained with only monocular 2D supervision. Here we show results on ShapeNet and FFHQ.
     }
    \label{fig:teaser}
\end{center}
}]

\newcommand\blfootnote[1]{%
  \begingroup
  \renewcommand\thefootnote{}\footnote{#1}%
  \addtocounter{footnote}{-1}%
  \endgroup
}

\blfootnote{
Project page: \url{https://github.com/Anciukevicius/RenderDiffusion}
\vspace{-10pt}
}

\begin{abstract}
\vspace{-2pt}
Diffusion models currently achieve state-of-the-art performance for both conditional and unconditional image generation.
However, so far, image diffusion models do not support tasks required for 3D understanding, such as view-consistent 3D generation or single-view object reconstruction. 
In this paper, we present \name, the first diffusion model for 3D generation and inference, trained using only monocular 2D supervision.
Central to our method is a novel image denoising architecture that generates and renders an intermediate three-dimensional
representation of a scene in each denoising step. 
This enforces a strong inductive structure within the diffusion process, providing a 3D consistent representation while only requiring 2D supervision. 
The resulting 3D representation can be rendered from any view.
We evaluate \name on FFHQ, AFHQ, ShapeNet and CLEVR datasets, showing competitive performance for generation of 3D scenes and inference of 3D scenes from 2D images. Additionally, our diffusion-based approach allows us to use 2D inpainting to edit 3D scenes.

\end{abstract}

\section{Introduction}

\cam{Image diffusion models} now
achieve state-of-the-art performance
on both generation and inference tasks. %
Compared to alternative approaches (e.g.~GANs and VAEs), they are able to model complex datasets more faithfully, particularly for long-tailed distributions, by explicitly maximizing likelihood of the training data. Many exciting applications have emerged in only the last few months, including text-to-image generation~\cite{ramesh2022hierarchical,saharia2022photorealistic}, inpainting~\cite{saharia22palette}, object insertion~\cite{avrahami2022blended},
and personalization~\cite{ruiz2022dreambooth}.

However, in \textit{3D} generation and understanding, their success has so far been limited, both in terms of quality and diversity of the results. Some methods have successfully applied diffusion models directly to point cloud or voxel data~\cite{luo2021diffusion,zhou20213d}, or optimized a NeRF using a pre-trained diffusion model~\cite{anonymous2023dreamfusion}.
This limited success in 3D is due to two problems: first, an explicit 3D representation (e.g., voxels) leads to significant memory demands and affects convergence speed; and more importantly, a setup that requires access to explicit 3D supervision is problematic as 3D model repositories contain orders of magnitude fewer data compared to image counterparts---a particular problem for large diffusion models which tend to be more data-hungry than GANs or VAEs.

In this work, we present \textit{\name} -- the first diffusion method for 3D content that is trained using only 2D images.
Like previous diffusion models, we train our model to denoise 2D images. 
Our key insight is to incorporate a latent 3D representation into the denoiser.
This creates an inductive bias that allows us to recover 3D objects while training only to denoise in 2D, \textit{without} explicit 3D supervision.
This latent 3D structure consists of a triplane representation \cite{eg3d} that is created from the noisy image by an encoder, and a volumetric renderer \cite{mildenhall2020nerf} that renders the 3D representation back into a (denoised) 2D image.
With the triplane representation, we avoid the cubic memory growth for volumetric data, and by working directly on 2D images, we avoid the need for 3D supervision.
Compared to latent diffusion models that work on a pre-trained latent space~\cite{rombach2021high, bautista2022gaudi}, working directly on 2D images also allows us to obtain sharper generation and inference results.
Note that \name does assume that we have the intrinsic and extrinsic camera parameters available at training time.

We evaluate \name on \cam{in-the-wild (FFHQ, AFHQ) and synthetic (CLEVR, ShapeNet) datasets} and show that it generates plausible and diverse 3D-consistent scenes (see Figure~\ref{fig:teaser}). Furthermore, we demonstrate that it successfully performs challenging inference tasks such as monocular 3D reconstruction and inpainting 3D scenes from masked 2D images, without specific training for those tasks. We show improved reconstruction accuracy over a state-of-the-art method~\cite{eg3d} in monocular 3D reconstruction that was also trained with only monocular supervision.

In short, our key contribution is a denoising architecture with an explicit latent 3D representation, which enables us to build \textbf{the first 3D-aware diffusion model that can be trained purely from 2D images}.

\section{Related Work}
\boldstart{Generative models.}
To achieve high-quality image synthesis, diverse generative models have been proposed, including GANs \cite{goodfellow2014generative,arjovsky2017wasserstein,karras2019style}, VAEs~\cite{kingma2013auto,van2017neural}, independent component estimation~\cite{dinh2014nice}, and autoregressive models~\cite{van2016pixel}.
Recently, diffusion models \cite{DBLP:journals/corr/Sohl-DicksteinW15,ho2020denoising} have achieved state-of-the-art results on image generation and many other image synthesis tasks \cite{kingma2021variational, dhariwal2021diffusion,ho2022cascaded,saharia2022image,lugmayr2022repaint}. 
Uniquely, diffusion models can avoid mode collapse (a common challenge for GANs), achieve better density estimation than other likelihood-based methods, and lead to high sample quality when generating images.  
We aim to extend such powerful image diffusion models from 2D image synthesis to 3D content generation and inference.

While many generative models (like GANs) have been extended for 3D generation tasks \cite{hologan,henzler2019escaping,nguyen2020blockgan}, applying diffusion models on 3D scenes is still relatively an unexplored area.
A few recent works build 3D diffusion models using point-, voxel-, or SDF-based representations \cite{zhou20213d,luo2021diffusion,hui2022neural,cheng23sdfusion,li22diffusionsdf,mueller22diffrf}, relying on  3D (geometry) supervision. 
Except for the concurrent works \cite{cheng23sdfusion,mueller22diffrf}, these methods focus on shape generation only and do not model surface color or texture --  which are important for rendering the resulting shapes.
Instead, we combine diffusion models with advanced neural field representations, leading to complete 3D content generation. Unlike implicit models \cite{watson2023novel}, our model generates both shape and appearance, allowing for realistic image synthesis under arbitrary viewpoints.

\boldstartspace{Neural field representations.}
There has been exponential progress in the computer vision community on representing 3D scenes as neural fields \cite{mildenhall2020nerf,barron2021mip,sun2022direct,xu2022point,muller2022instant,chen2022tensorf}, allowing for high-fidelity rendering in various reconstruction and image synthesis tasks \cite{park2021nerfies,li2021neural,peng2021neural,athar2022rignerf,zhang2022arf}.
We utilize the recent tri-plane based representation \cite{eg3d,chen2022tensorf} in our diffusion model, allowing for compact and efficient 3D modeling with volume rendering.

Recent NeRF-based methods have developed generalizable networks \cite{yu2021pixelnerf,chen2021mvsnerf} trained across scenes for efficient few-shot 3D reconstruction. 
While our model is trained only on single-view images, our method can achieve high inference quality comparable to such a method like PixelNeRF \cite{yu2021pixelnerf} that requires multi-view data during training. 
Several concurrent approaches use 2D diffusion models as priors for this task \cite{zhou23sparsefusion,deng22nerdi,gu23nerfdiff}; unlike ours they cannot synthesise new images or scenes \textit{a priori}.

Neural field representations have also been extended to building 3D generative models \cite{chan2021pi,kosiorek2021nerf}, where most methods are based on GANs \cite{eg3d,gu2021stylenerf,or2022stylesdf, gao2022get3d}. 
Two concurrent works -- DreamFusion \cite{anonymous2023dreamfusion} (extending DreamFields \cite{jain2021dreamfields}) and Latent-NeRF \cite{metzer22latentnerf} -- leverage pre-trained 2D diffusion models as priors to drive NeRF optimization, achieving promising text-driven 3D generation.

We instead seek to build a 3D diffusion model and achieve direct object generation via sampling.
Another concurrent work, GAUDI \cite{bautista2022gaudi}, introduces a 3D generative model 
that first learns a triplane-based latent space using multi-view data, and then builds a diffusion model over this latent space. In contrast, our approach only requires single-view 2D images and enables end-to-end 3D generation from image diffusion without pre-training any 3D latent space. %
DiffDreamer \cite{cai2022diffdreamer} casts 3D scene generation as repeated inpainting of RGB-D images rendered from a moving camera; this inpainting uses a 2D diffusion model.
However this method cannot generate scenes \textit{a priori}.

\begin{figure}[t]
  \centering
\includegraphics[width=\linewidth]{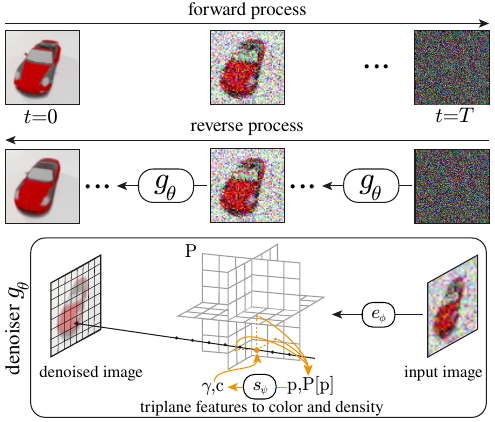}
  \caption{
  \textbf{Architecture overview. } 
  Images are generated by iteratively applying the denoiser $g_\theta$ to noisy input images, progressively removing the noise. Unlike traditional 2D diffusion models, our denoiser contains 3D structure in the form of a triplane representation $\mathbf{P}$ that is inferred from a noisy input image by the encoder $e_\phi$. A small MLP $s_\psi$ converts triplane features at arbitrary sample points into colors and densities that can then be rendered back into a denoised output image using a volumetric renderer.}
  \label{fig:arch}
  \vspace{-1em}
\end{figure}

\section{Method}
Our method builds on the successful training and generation setup of 2D image diffusion models, which are trained to denoise input images that have various amounts of added noise~\cite{ho2020denoising}. At test time, novel images are generated by applying the model in multiple steps to progressively recover an image starting from pure noise samples. We keep this training and generation setup, but modify the architecture of the main denoiser to encode the noisy input image into a 3D representation of the scene 
that is volumetrically rendered to obtain the denoised output image.
This introduces an inductive bias that favors 3D scene consistency, and allows us to render the 3D representation from novel viewpoints. Figure~\ref{fig:arch} shows an overview of our architecture.
In the following, we first briefly review 2D image diffusion models (Section~\ref{sec:image_diffusion}), then describe the novel architectural changes we introduce to obtain a 3D-aware denoiser (Section~\ref{sec:3d_aware_denoiser}).

\subsection{Image Diffusion Models}
\label{sec:image_diffusion}

Diffusion models generate an image $\mathbf{x}_0$ by moving a starting image $\mathbf{x}_T \sim \mathcal{N}(\mathbf{0}, \mathbf{I})$
progressively closer to the data distribution through multiple denoising steps $\mathbf{x}_{T-1}, \dots, \mathbf{x}_0$.

\paragraph{Forward process.} To train the model, noisy images $\mathbf{x}_1, \dots, \mathbf{x}_T$ are created by repeatedly adding Gaussian noise starting from a training image $\mathbf{x}_0$:
\begin{equation}
    q(\mathbf{x}_t | \mathbf{x}_{t-1}) \sim \mathcal{N}(\mathbf{x}_t; \sqrt{1-\beta_t}\mathbf{x}_{t-1}, \beta_t \mathbf{I}),
\end{equation}    
where $\beta_t$ is a variance schedule that increases from $\beta_0=0$ to $\beta_T=1$ and controls how much noise is added in each step. We use a cosine schedule~\cite{nichol2021improved} in our experiments. To avoid unnecessary iterations, we can directly obtain 
$\mathbf{x}_t$ from $\mathbf{x}_0$ in a single step using the closed form:
\begin{gather}
     q(\mathbf{x}_t | \mathbf{x}_0) = \sqrt{\bar{\alpha_t}}\mathbf{x}_{0} + \sqrt{1-\bar{\alpha_t}}\boldsymbol{\epsilon} \nonumber\\
     \text{with } \boldsymbol{\epsilon} \sim \mathcal{N}(0, \mathbf{I}), \bar{\alpha_t} \coloneqq \prod_{s=1}^{t} \alpha_s \text{ and } \alpha_t \coloneqq (1-\beta_t). 
     \label{eq:forward_noise}
\end{gather}
\paragraph{Reverse process.} The reverse process aims at reversing the steps of the forward process by finding the posterior distribution for the less noisy image $\mathbf{x}_{t-1}$ given the more noisy image $\mathbf{x}_t$:
\begin{align}
    q(\mathbf{x}_{t-1} | \mathbf{x}_t, \mathbf{x}_0) & \sim \mathcal{N}(\mathbf{x}_{t-1}; \boldsymbol{\mu_t}, \sigma_t^2\mathbf{I}), \\
    \text{where~ } \boldsymbol{\mu_t} &\coloneqq \frac{\sqrt{\bar{\alpha}_{t-1}}\beta_t}{1-\bar{\alpha}_{t}} \mathbf{x}_0 + \frac{\sqrt{\alpha_t}(1-\bar{\alpha}_{t-1})}{1-\bar{\alpha}_{t}}\mathbf{x}_t \nonumber \\
    \text{and~ } \sigma_t^2 &\coloneqq \frac{1-\bar{\alpha}_{t-1}}{1-\bar{\alpha}_t}\beta_t. \nonumber 
\end{align}
Note that $\mathbf{x}_0$ is unknown (it is the image we want to generate), so we cannot directly compute this distribution, instead we train a denoiser $g_\theta$ with parameters $\theta$ to approximate it. Typically only the mean $\boldsymbol{\mu}_t$ needs to be approximated by the denoiser, as the variance does not depend on the unknown image $\mathbf{x}_0$. Please see Ho et al.~\cite{ho2020denoising} for a derivation.

We could directly train a denoiser to predict the mean $\boldsymbol{\mu}_t$, however, Ho et al.~\cite{ho2020denoising} show that a denoiser $g_\theta$ can be trained more stably and efficiently by directly predicting the total noise $\boldsymbol{\epsilon}$ that was added to the original image $\mathbf{x}_0$ in Eq.~\ref{eq:forward_noise}. We follow Ho et al., but since our denoiser $g_\theta$ will also be tasked with reconstructing a 3D version of the scene shown in $\mathbf{x}_0$ as intermediate representation, we train $g_\theta$ to predict $\mathbf{x}_0$ instead of the noise $\boldsymbol{\epsilon}$:
\begin{equation}
    L \coloneqq \| g_\theta(\mathbf{x}_t, t) - \mathbf{x}_0 \|_1,
\end{equation}
where $L$ denotes the training loss.
Once trained, at generation time, the model $g_\theta$ can then approximate the mean $\boldsymbol{\mu_t}$ of the posterior $q(\mathbf{x}_{t-1} | \mathbf{x}_t, \mathbf{x}_0) = \mathcal{N}(\mathbf{x}_{t-1}; \boldsymbol{\mu_t}, \sigma_t^2\mathbf{I})$ as:
\begin{equation}
    \boldsymbol{\mu}_t \approx \frac{1}{\sqrt{\alpha_t}} \left(\mathbf{x}_t - \frac{1-\alpha_t}{1-\bar{\alpha}_t}\big(\mathbf{x}_t - \sqrt{\bar{\alpha_t}} g_\theta(\mathbf{x}_t, t)\big)\right).
\end{equation}
This approximate posterior is sampled in each generation step to progressively get the less noisy image $\mathbf{x}_{t-1}$ from the more noisy image $\mathbf{x}_t$.

\subsection{3D-Aware Denoiser}
\label{sec:3d_aware_denoiser}
The denoiser $g_\theta$ takes a noisy image $\mathbf{x}_t$ as input and outputs a denoised image $\tilde{\mathbf{x}}_0$. In existing methods~\cite{ho2020denoising,Rombach_2022_CVPR}, the denoiser $g_\theta$ is typically implemented by a type of UNet~\cite{RonnebergerFB15}. This works well in the 2D setting but does not encourage the denoiser to reason about the 3D structure of a scene. 
We introduce a latent 3D representation into the denoiser based on \emph{triplanes}~\cite{eg3d, peng2020convolutional}. For this purpose, we modify the architecture of the denoiser to incorporate two additional components: a \emph{triplane encoder} $e_\phi$ that transforms the input image $\mathbf{x}_t$, posed using camera view $\mathbf{v}$, into a 3D triplane representation, and a \emph{triplane renderer} $r_\psi$ that renders the 3D triplane representation back into a denoised image $\tilde{\mathbf{x}}_0$, such that, 
\begin{equation}
    g_\theta(\mathbf{x}_t,t,\mathbf{v}) \coloneqq r_\psi(e_\phi(\mathbf{x}_t, t), \mathbf{v}),
\end{equation}
where $\theta$ denotes the concatenated parameters $\phi$ and $\psi$ of the encoder and renderer.
The output image is a denoised version of the input image, thus it has to be rendered from the same viewpoint. We assume the viewpoint $\mathbf{v}$ of the input image to be available. Note that the noise is applied directly to the source/rendered images.

\vspace{-10pt}
\paragraph{Triplane representation.}
A triplane representation
$\mathbf{P}$
factorizes a full 3D feature grid into three 2D feature maps placed along the three (canonical) coordinate planes, giving a significantly more compact representation~\cite{peng2020convolutional, eg3d}. Each feature map has a resolution of $N \times N \times n_f$, where $n_f$ is the number of feature channels. The feature for any given 3D point $\mathbf{p}$ is then obtained by projecting the point to each coordinate plane, interpolating each feature map bilinearly, and summing up the three results to get a single feature vector of size $n_f$. We denote this process of bilinear sampling from $\mathbf{P}$ as $\mathbf{P}[\mathbf{p}]$.
\vspace{-10pt}
\paragraph{Triplane encoder.} The triplane encoder $e_\phi$ transforms an input image $\mathbf{x}_t$ of size $M \times M \times 3$ into a triplane representation of size $N \times N \times 3n_f$, where $N \ge M$. We use the U-Net architecture commonly employed in diffusion models \cite{ho2020denoising} as a basis, but append additional layers (without skip connections) to output feature maps in the size of the triplanes. More architectural details are given in the supplementary material.

\vspace{-10pt}
\paragraph{Triplane renderer.} The triplane renderer $r_\psi$ performs volume rendering using the triplane features and outputs an image $\mathbf{x}_{t-1}$ of size $M \times M \times 3$. 
At each 3D sample point $\mathbf{p}$ along rays cast from the image, we obtain density $\gamma$ and color $\mathbf{c}$ with an MLP as $(\gamma, \mathbf{c}) = s_\psi(\mathbf{p}, \mathbf{P}[\mathbf{p}])$. The final color for a pixel is produced by integrating colors and densities along a ray using the same explicit volume rendering approach as MipNeRF~\cite{barron2021mip}.
We use the two-pass importance sampling approach of NeRF~\cite{mildenhall2020nerf}; the first pass uses stratified sampling to place samples along each ray, and the second pass importance-samples the results of the first pass.

\subsection{Score-Distillation Regularization}
\cam{
To avoid solutions with trivial geometry on FFHQ and AFHQ, we found that it is helpful to regularize the model with a score distillation loss \cite{anonymous2023dreamfusion}.
This encourages the model to output a scene that looks plausible from a random viewpoint $\mathbf{v}_r$ sampled from the training set, not just the viewpoint $\mathbf{v}$ of the input image $\mathbf{x}_t$.
Specifically, at each training step, we render the denoised 3D scene $e_\phi(\mathbf{x}_t, t)$ from $\mathbf{v}_r$, giving the image $\tilde{\mathbf{x}}_r$ and compute a score distillation loss for $\tilde{\mathbf{x}}_r$ as:
$\|\tilde{\mathbf{x}}_r - g_\theta(\sqrt{\bar{\alpha_t}}\tilde{\mathbf{x}}_r + \sqrt{1-\bar{\alpha_t}} \boldsymbol{\epsilon}, t, \mathbf{v}_r)\|_1$
with $\boldsymbol{\epsilon} \sim \mathcal{N}(0, \mathbf{I})$.
}

\subsection{3D Reconstruction}
Unlike existing 2D diffusion models, we can use \name to reconstruct 3D scenes from 2D images. To reconstruct the scene shown in an input image $\mathbf{x}_0$, we pass it through the forward process for $t_r \le T$ steps, and then denoise it in the reverse process using our learned denoiser $g_\theta$. In the final denoising step, the triplanes encode a 3D scene that can be rendered from novel viewpoints. The choice of $t_r$ introduces an interesting control that is not available in existing 3D reconstruction methods. It allows us to trade off between reconstruction fidelity and generalization to out-of-distribution input images: At $t_r=0$, no noise is added to the input image and the 3D reconstruction reproduces the scene shown in the input as accurately as possible; however, out-of-distribution images cannot be handled. With larger values for $t_r$, input images that are increasingly out-of-distribution can be handled, as the denoiser can move the input images towards the learned distribution. This comes at the cost of reduced reconstruction fidelity, as the added noise removes some detail from the input image, which the denoiser fills in with generated content.

\begin{table*}
\centering
\caption{\small 
\textbf{3D reconstruction performance.} We compare to EG3D~\cite{eg3d} and PixelNeRF~\cite{yu2021pixelnerf} on ShapeNet and our variant of the CLEVR dataset (CLEVR1). Since PixelNeRF has a significant advantage due to training with multi-view supervision, we keep it in a separate category and denote with bold numbers the best among the two methods with single-view supervision, i.e.~EG3D and \name.}
\label{tab:reconstruction}
\scalebox{0.9}{
\begin{tabular}{@{}rcccccccccc@{}}
\toprule  
& \multicolumn{8}{c}{ShapeNet} & \multicolumn{2}{c}{CLEVR1} \\
\cmidrule(lr){2-9} \cmidrule(l){10-11}
& \multicolumn{2}{c}{\texttt{car}} & \multicolumn{2}{c}{\texttt{plane}} & \multicolumn{2}{c}{\texttt{chair}} & \multicolumn{2}{c}{average} \\
\cmidrule(lr){2-3} \cmidrule(lr){4-5} \cmidrule(lr){6-7} \cmidrule(lr){8-9}
 & PSNR & SSIM & PSNR & SSIM & PSNR & SSIM & PSNR & SSIM & PSNR & SSIM\\
\midrule
PixelNeRF~\cite{yu2021pixelnerf} & 27.3 & 0.838  & 29.4 & 0.887  & 30.2 & 0.884  & 28.9 & 0.870  & 43.4 & 0.988\\
\midrule
EG3D~\cite{eg3d} & 21.8 & 0.714  & 25.0 & 0.803  & 25.5 & 0.803  & 24.1 & 0.773  & 33.2 & 0.910  \\
\name (ours) & \textbf{25.4} & \textbf{0.805}  & \textbf{26.3} & \textbf{0.834}  & \textbf{26.6} & \textbf{0.830}  & \textbf{26.1} & \textbf{0.823}  & \textbf{39.8} & \textbf{0.976}  \\
 \bottomrule   
\end{tabular}
}
\end{table*}

\begin{figure}[b!]
\centering
\includegraphics[width=\linewidth]{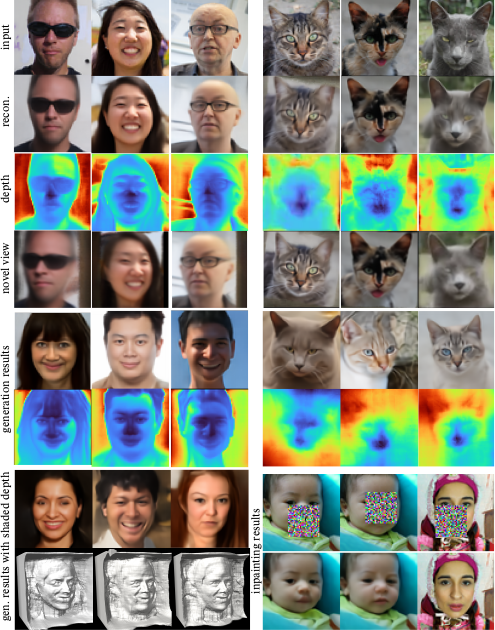}
\vspace{-20pt}
\caption{\textbf{\name results on FFHQ and AFHQ.} We show reconstruction (top four rows), unconditional generation (bottom left), and 3D-aware inpainting (bottom right).}
\label{fig:generations-fhq}
\vspace{-15pt}
\end{figure}

\section{Experiments}

We evaluate \name on three tasks: monocular 3D reconstruction, unconditional generation, and 3D-aware inpainting.

\begin{figure*}[t]
  \centering
\includegraphics[width=\linewidth]{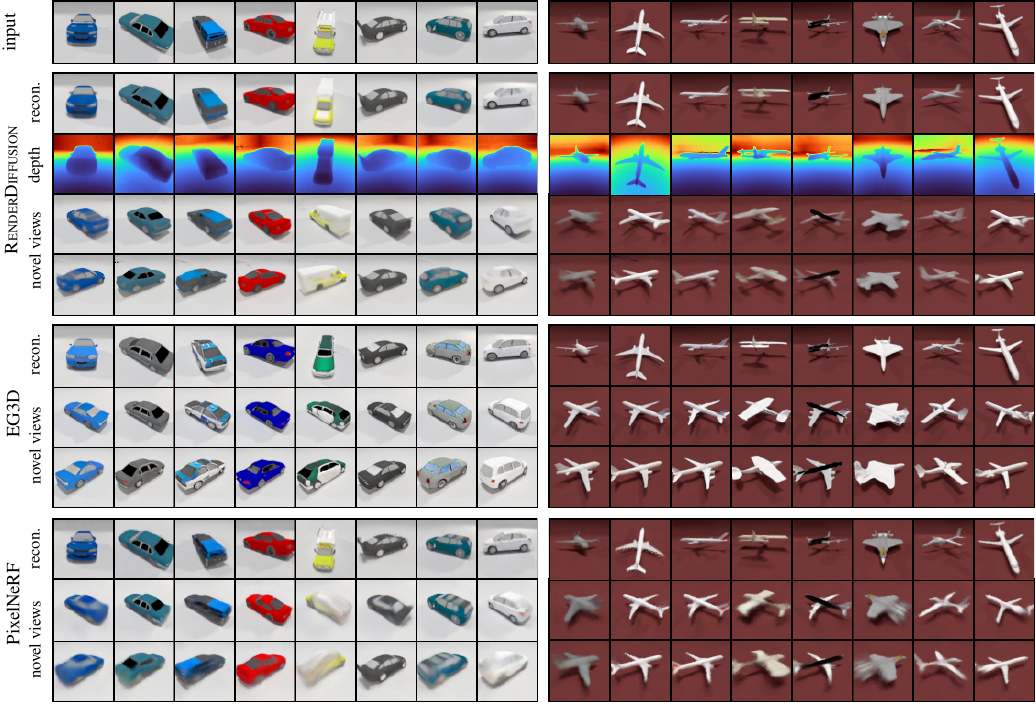}
\vspace{-15pt}
  \caption{
  \textbf{Reconstruction quality.}  We compare our results on ShapeNet \texttt{car} and \texttt{plane}  to PixelNeRF and EG3D (through inversion). Compared to EG3D, our reconstructions better preserve shape identity; compared to PixelNeRF, ours are sharper and more detailed.}
  \label{fig:reconstruction}
  \vspace{-1.1em}
\end{figure*}

\vspace{-1em}
\paragraph{Datasets.}
\cam{For training and evaluation we use real-world human face dataset (FFHQ), a cat face dataset (AFHQv2) as well as generated datasets of scenes from CLEVR~\cite{johnson17clevr} and ShapeNet~\cite{shapenet2015}.
We adopt FFHQ and AFHQv2 from EG3D \cite{eg3d}, which uses off-the-shelf estimator to extract approximate extrinsics and augments dataset with horizontal image flips.
}
We generate a variant of the CLEVR dataset which we call the \emph{CLEVR1} dataset, where each scene contains a single object standing on the plane at the origin. We randomize the objects in different scenes to have different colors, shapes, and sizes and generate $900$ scenes, where $400$ are used for training and the rest for testing. To evaluate our method on more complex shapes, we use objects from three categories of the ShapeNet dataset. ShapeNet contains man-made objects of various categories that are represented as textured meshes. We use shapes from the \texttt{car}, \texttt{plane}, and \texttt{chair} categories, each placed on a ground-plane. We use a total of $3200$ objects from each category: $2700$ for training and $500$ for testing. To render the scenes, we sample $100$ viewpoints uniformly on a hemisphere centered at the origin. %
Viewing angles that are too shallow (below $12^\circ$) are re-sampled. $70$ of these viewpoints are used for training, and $30$ are reserved for testing. We use Blender~\cite{blender} to render each of the objects from each of the $100$ viewpoints. %

\subsection{Monocular 3D Reconstruction}
We evaluate 3D reconstruction on test scenes from each of our three ShapeNet categories. %
Since these images are drawn from the same distribution as the training data, we do not add noise, i.e. we set $t_r=0$. Reconstruction is performed on the non-noisy image with one iteration of our denoiser $g_\theta$. Note that our method does not require the camera viewpoint of the image as input.

\vspace{-1.3em}
\paragraph{Baselines.}
We compare to two state-of-the-art methods %
that are also trained without 3D supervision. \emph{EG3D}~\cite{eg3d} is a generative model that uses triplanes as its 3D representation and, like our method, trains with only single images as supervision. As it does not have an encoder, we perform GAN inversion to obtain a 3D reconstruction~\cite{zhu16manipulation}. 
Specifically, we optimize the latent vector $z$ and latent noise maps $n$ of the StyleGAN2~\cite{karras2020analyzing} decoder and super-resolution blocks in EG3D to match the input image when rendered from the ground truth viewpoint. We optimize EG3D for 1000 steps for each of $4$ random initializations, minimizing the difference in VGG16 features~\cite{simonyan15vgg} between generated and target images, then pick the best result.
As a second baseline we use \emph{PixelNeRF}~\cite{yu2021pixelnerf}, a recent approach for novel view synthesis from sparse views. Unlike our method, this is trained with multi-view supervision, giving it a significant advantage over both ours and EG3D. We therefore treat this baseline as a reference that we do not expect to outperform.
Both EG3D and PixelNeRF were carefully tuned and re-trained on our datasets.

\vspace{-1.3em}
\paragraph{Metrics.}
We evaluate the 3D reconstruction performance by comparing all held-out test set views of the reconstructed scenes to the ground truth renders using two metrics: PSNR and SSIM~\cite{wang2004image}. We average the results over all test images.
Evaluating in image-space from multiple views has the advantage that it measures the accuracy not only of the shape, but also of the (possibly view-dependent) color of the surface at every point.

\begin{figure*}[t]
  \centering
\includegraphics[width=\linewidth]{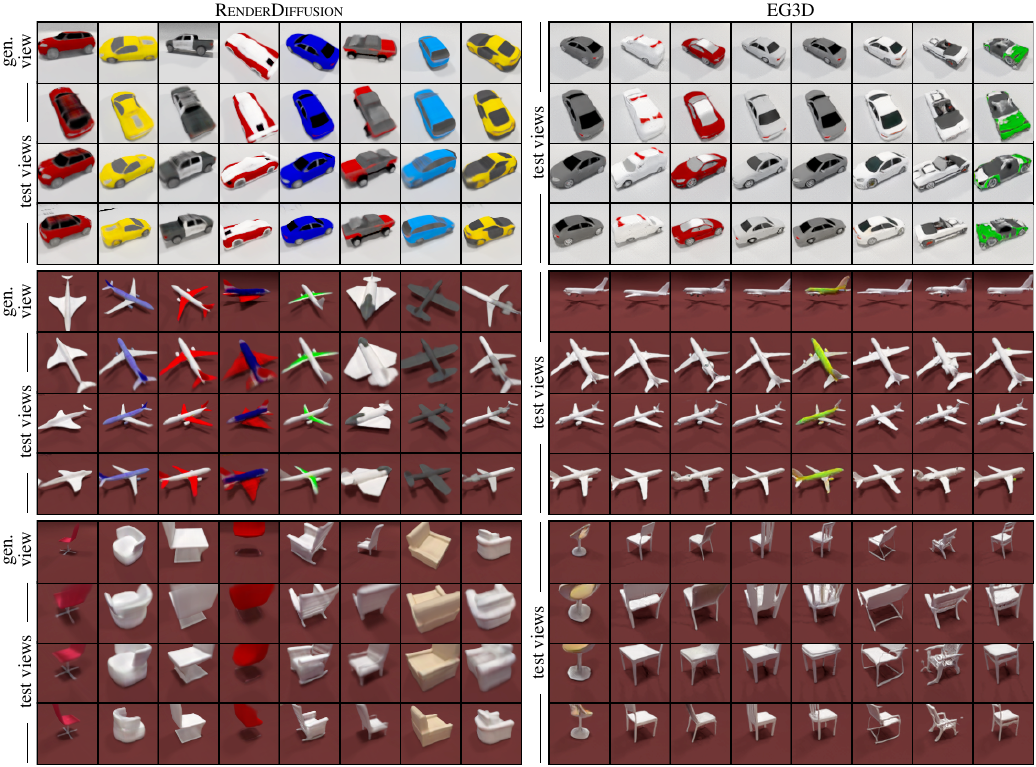}
\vspace{-15pt}
  \caption{
  \textbf{Unconditional generation.} 
  Results from \name and EG3D on ShapeNet categories; for \name, we show the view used during the reverse process in the first row. Note how our scenes have competitive quality and diversity compared to EG3D.}
  \label{fig:generation}
  \vspace{-1em}
\end{figure*}

\begin{table}[b!]
    \centering
    \caption{
\textbf{Quantitative evaluation. } 
Generation results (coverage, higher is better) on ShapeNet, FFHQ, and AFHQ for \name, EG3D, and GIRAFFE, respectively. Other metrics are given in the supplementary.}
    \scalebox{0.93}{
    \begin{tabular}{@{}lccccc@{}}
    \toprule
        & \multicolumn{3}{c}{ShapeNet} & FFHQ & AFHQ \\
    \cmidrule(r){2-4}\cmidrule(lr){5-5}\cmidrule(l){6-6}
        & \texttt{car} & \texttt{plane} & \texttt{chair} \\
    \midrule
        GIRAFFE \cite{niemeyer2021giraffe} & 0.11 & 0.45 & 0.39 & 0.66 & 0.07 \\
        EG3D \cite{eg3d} & \textbf{0.70} & 0.61 & 0.71 & \textbf{0.68} & \textbf{0.37} \\
        Ours & 0.46 & \textbf{0.84} & \textbf{0.85} & 0.31 & 0.26 \\
    \bottomrule
    \end{tabular}
    }
    \label{tab:coverage}
    \vspace{-1.1em}
\end{table}

\vspace{-15pt}
\paragraph{Results.}
Qualitative results from our method and the baselines on ShapeNet
are shown in Figure~\ref{fig:reconstruction} and additional results including the CLEVR dataset, are shown in the supp. material.
We see that EG3D usually predicts shapes that appear realistic from all angles.
However, these often differ somewhat in shape or color from the input image, sometimes drastically (e.g. the 4\textsuperscript{th} and 5\textsuperscript{th} cars).
In contrast, PixelNeRF produces faithful reconstructions of its input views.
However, when rotated away from an input view the images are often blurry or lack details.
Our \name achieves a balance between the two, preserving the identity of shapes, but yielding sharp and plausible details in back-facing regions.
Quantitative results are shown in Table~\ref{tab:reconstruction}. In agreement with the qualitative results, all methods perform better on the simpler CLEVR1 dataset than on the three ShapeNet classes.
\name out-performs EG3D across all datasets, achieving an average PSNR on ShapeNet of 26.1, versus 24.1 for EG3D.
PixelNeRF, which is not directly comparable since it receives multi-view supervision during training, achieves higher performance still, with an average PSNR on ShapeNet of 28.9. \cam{Note that on GeForce GTX 1080 Ti our method performs reconstruction in a single pass in under $0.03$ seconds per scene compared to $\sim$3 min. for EG3D inversion.}
\cam{On FFHQ and AFHQ (\fig{generations-fhq}, top four rows), our method accurately reconstructs input faces and cats, predicting a plausible depth map and renderings from novel views.}

\begin{figure}[t]
  \centering
    \includegraphics[width=\linewidth]{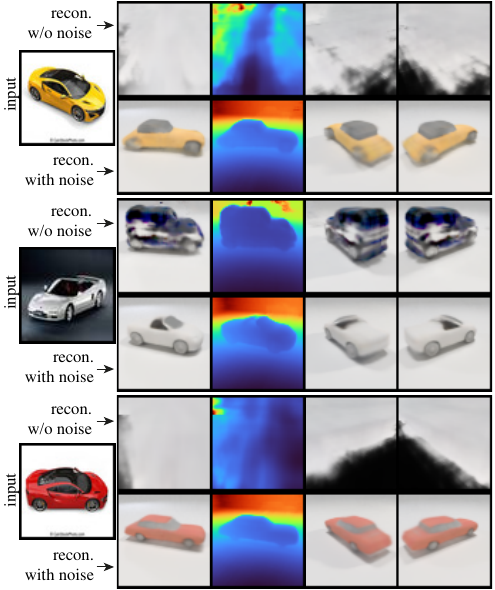}
  \caption{\textbf{Reconstruction of out-of-distribution images.} By adding noise to input images and reconstructing with multiple steps of the reverse process, we trade off between reconstruction fidelity and generalization to out-of-distribution (OOD) images. %
  }
  \label{fig:recon_ood}
  \vspace{-1,5em}
\end{figure}

\vspace{-15pt}
\paragraph{Reconstruction on out-of-distribution images.}
Using a 3D-aware denoiser allows us to reconstruct a 3D scene from noisy images, where information that is lost to the noise is filled in with generated content. By adding more noise, we can generalize to input images that are increasingly out-of-distribution, at the cost of reconstruction fidelity. In Figure~\ref{fig:recon_ood}, we show 3D reconstructions from photos that have significantly different backgrounds and materials than the images seen at training time. We see that results with added noise ($t_r = 40$) generalize better than results without added noise ($t_r=0$), at the cost of less accurate shapes and poses of the reconstructed models.
In addition, in the supplementary material, we show how results vary with differing amounts of noise $t_r$.

\subsection{Unconditional Generation}
We show results for unconditional generation; for additional quantitative results, please refer to the supplemental.

\vspace{-10pt}
\paragraph{Baselines.}
We compare against EG3D~\cite{eg3d}, which is the most similar existing work to ours, since it too uses a triplane representation for the 3D scene, and a similar rendering approach. \cam{We also compare with the older methods pi-GAN \cite{chan2021pi} and GIRAFFE \cite{niemeyer2021giraffe}. All three baselines use adversarial training rather than denoising diffusion.}

\paragraph{Results.} %
Qualitative examples from both methods are shown in \fig{generation}.
We see that scenes generated by both models appear realistic from all viewpoints, and are 3D-consistent.
This is particularly notable for \name, since our method performs the denoising process from just a single viewpoint (top rows of \fig{generation}).
We observe somewhat higher diversity of both color and shape among the samples from our model than those from EG3D.
However, both methods are able to generate complex structures (e.g. slatted chair backs), and synthesise physically-plausible shadows cast onto the ground-plane.
\cam{Quantitatively (\tab{coverage}) our method performs better than EG3D and GIRAFFE w.r.t. dataset coverage (see supp. for a definition and additional results) on two of the three ShapeNet classes, while EG3D is best for the other datasets.}
\cam{On FFHQ and AFHQ, our model again produces plausible, 3D-consistent samples (\fig{generations-fhq}), though with some artifacts visible at larger azimuth angles.}

\subsection{3D-Aware Inpainting}

\cam{We also apply our} trained model to the task of inpainting masked 2D regions of an image while simultaneously reconstructing the 3D shape it shows.
We follow an approach similar to RePaint~\cite{lugmayr2022repaint}, but using our 3D denoiser instead of their 2D architecture.
Specifically, we condition the denoising iterations on the known regions of the image, 
by setting $\mathbf{x}_{t-1}$ in known regions to the noised target pixels, while sampling the unknown regions as usual based on $\mathbf{x}_t$.
Thus, the model performs 3D-aware inpainting, finding a latent 3D structure that is consistent with the observed part of the image, and also plausible in the masked part.

\begin{figure}[t]
  \centering
    \includegraphics[width=\linewidth]{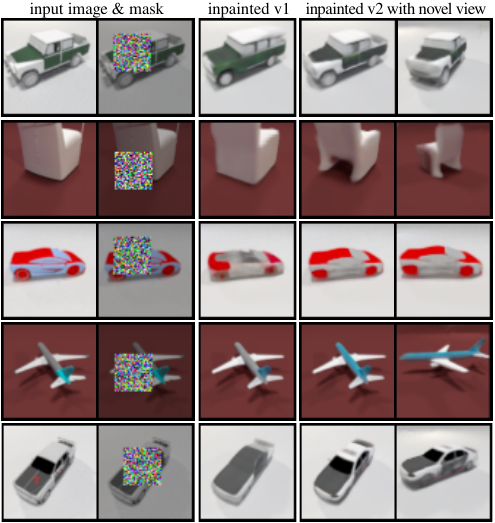}
  \caption{
  \textbf{3D-aware inpainting.} We inpaint the noisy 2D region shown on the left starting from two different noise samples, resulting in two different inptained 3D scenes. We also show a novel view for the second scene.}
  \label{fig:inpainting}
  \vspace{-1,5em}
\end{figure}

\paragraph{Results.}
Qualitative results are shown in Figure~\ref{fig:inpainting}. For each masked image, we show scnes resulting from two different noise samples. Even though our model was not trained explicitly on this task, we can see that it generates diverse and plausible inpainted regions that match the observed part of the image.
In the supplementary material, we give quantitative results on this task, by sampling multiple completions for each input, and measuring how close they can get to the ground-truth.

\section{Conclusion}

We have presented \name, the first 3D diffusion model that can be trained purely from posed 2D images, without requiring any explicit 3D supervision.
Our denoising architecture incorporates a triplane rendering model which enforces a strong inductive bias and produces 3D-consistent generations. 
\name can be used to infer a 3D scene from an image, for 3D editing using 2D inpainting, and for 3D scene generation. We have shown competitive performance on sampling and inference tasks, in terms of both quality and diversity of results.

Our method currently has several limitations. First, our generated images still lag behind GANs in some cases, probably due to more blurry outputs; however ours is the first 3D-aware diffusion model trained with 2D images, and there are engineering techniques that could enhance the quality, like upsampling models.
Second, we have introduced a score distillation regularization to prevent learning trivial geometry on FFHQ/AFHQ, which results in a loss of fidelity in the generated 3D models; we expect this to be less of an issue as score distillation methods improve.
Third, we require the training images to include camera extrinsics, and our triplanes are positioned in a global coordinate system, which restricts generalization across object placements. This limitation can be addressed by utilizing off-the-shelf pose estimation, or rendering everything in the camera-view.
Finally, we would like to support object editing and material editing, enabling a more expressive 3D-aware 2D image-editing workflow.

\section{Acknowledgements}

We would like to thank 
Yilun Du,
Christopher K. I. Williams,
Noam Aigerman,
Julien Philip 
and 
Valentin Deschaintre
for valuable discussions.
HB was supported by the EPSRC Visual AI grant EP/T028572/1; NM was partially supported by the UCL AI Centre. 

{\small
\bibliographystyle{ieee_fullname}
\bibliography{bibliography}
}

\clearpage

\renewcommand{\thesection}{S\arabic{section}}
\renewcommand{\thetable}{S\arabic{table}}
\renewcommand{\thefigure}{S\arabic{figure}}

\section{Overview of Supplementary}
\label{sec:overview}

In this supplementary material, we provide additional architecture details (Section~\ref{sec:supp_arch}), additional results for unconditional generation (Section~\ref{sec:supp_unconditional}), additional results for 3D-aware inpainting (Section~\ref{sec:supp_inpainting}), an additional experiment where generate multiple ShapeNet categories with a single model (Section~\ref{sec:supp_multimodal}), and additional results for reconstruction (Section~\ref{sec:supp_oodinference}).

\section{Architecture Details}

Here we summarise the architecture of the denoiser network. Code, training configurations and datasets are publicly available at \url{https://github.com/Anciukevicius/RenderDiffusion}.

\label{sec:supp_arch}
\paragraph{Triplane encoder.}
The triplane encoder transforms the input image of size $M \times M \times 3$ into a triplane representation of size $N \times N \times 3n_f$. We choose $M=64$ and $N=256$ for our experiments on ShapeNet, as we found that the increased triplane resolution improves the quality of our results, and $M=32$, $N=32$ for CLEVR1. Similar to other 2D diffusion models~\cite{ho2020denoising}, we use a UNet~\cite{RonnebergerFB15} architecture for the triplane encoder. The UNet consists of 8 down and up blocks \cite{RonnebergerFB15}. Each block consists of 2 ResNet blocks \cite{He2016DeepRL} that additionaly take a timestep embedding, and linear attention. 
If the triplane has larger resolution than the input image, we append additional up blocks to the UNet that upsample the image to the triplane resolution. These have the same architecture as the other UNet blocks, except that they do not use skip connections, as there is no down block in the UNet with the corresponding resolution.

\paragraph{Triplane renderer.}

To render triplanes, we mostly follow EG3D~\cite{eg3d}; however, we use explicit volumetric rendering that samples points along the ray and queries a 2-layer fully-connected neural network to output color and a density \cite{eg3d}. The network takes as input 32-dimensional sum-pooled interpolations of triplane features. Unlike EG3D, we also use a positional embedding of the 3D sample position \cite{peng2020convolutional} as input to the network, which allows the network to represent parts of the ground plane that extend beyond the triplanes with a single constant feature.

\begin{figure}[t]
  \centering
    \includegraphics[width=\linewidth]{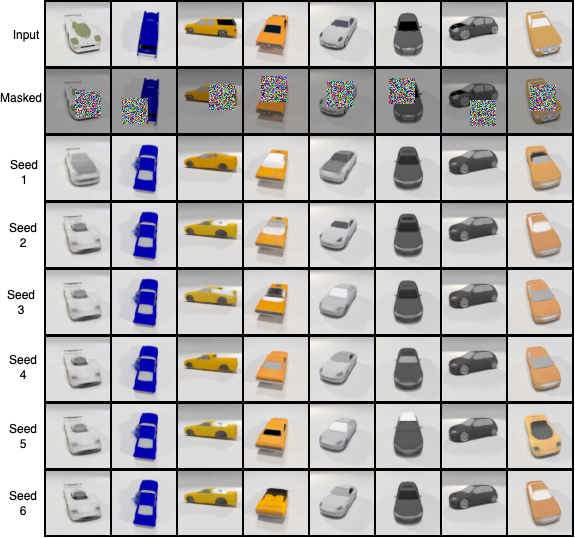}
  \caption{3D reconstructions over $6$ random seeds given masked input image from \texttt{car} test set. Notice the diversity in predictions. Results were selected randomly (non-cherry-picked). Since there are many possible inpaintings, \tab{supp_inpainting_table} reports quantitative results by taking best-performing reconstruction.}
  \label{fig:inpainting}
  \vspace{-1,5em}
\end{figure}

\begin{table*}[t]
\centering
\caption{\small 3D generation performance for our model RenderDiffusion, and baselines GIRAFFE~\cite{niemeyer2021giraffe}, pi-GAN~\cite{chan2021pi} and EG3D~\cite{eg3d}, on ShapeNet and CLEVR1 datasets. We report FID for train viewpoints for all methods, and also for test viewpoints with ours and EG3D, as well as coverage (\textit{cov.}) and density (\textit{dens.}) \cite{naeem2020reliable}}
\label{tab:generation}
\vspace{-8pt}
\scalebox{0.93}{%
\setlength{\tabcolsep}{2pt}%
\begin{tabular}{@{}lcccccccccccccccccccc@{}}
\toprule  
& \multicolumn{16}{c}{ShapeNet} & \multicolumn{4}{c}{CLEVR1} \\
\cmidrule(lr){2-17} \cmidrule(l){18-21}
& \multicolumn{4}{c}{\texttt{car}} & \multicolumn{4}{c}{\texttt{plane}} & \multicolumn{4}{c}{\texttt{chair}} & \multicolumn{4}{c}{average} \\
\cmidrule(lr){2-5} \cmidrule(lr){6-9} \cmidrule(lr){10-13} \cmidrule(lr){14-17}
 & $\text{FID}_r$ & $\text{FID}_t$ & $\text{cov.}$ & $\text{dens.}$ & $\text{FID}_r$ & $\text{FID}_t$ & $\text{cov.}$ & $\text{dens.}$ & $\text{FID}_r$ & $\text{FID}_t$ & $\text{cov.}$ & $\text{dens.}$ & $\text{FID}_r$ & $\text{FID}_t$ & $\text{cov.}$ & $\text{dens.}$ & $\text{FID}_r$ & $\text{FID}_t$ & $\text{cov.}$ & $\text{dens.}$ \\
\midrule
GIRAFFE~\cite{niemeyer2021giraffe} & 
30.5 & -- & 0.11 & 0.03 & %
56.1 & -- & 0.45 & 0.41 & %
35.2 & -- & 0.39 & 0.30 & %
40.6 & -- & 0.32 & 0.25 & %
-- & -- & -- & -- \\ %
pi-GAN~\cite{chan2021pi} & 
25.6 & -- & 0.07 & 0.02 & %
33.5 & -- & 0.16 & 0.08 & %
41.4 & -- & 0.14 & 0.05 & %
33.5 & -- & 0.12 & 0.15 & %
36.0 & -- & 0.04 & 0.002 \\ %
EG3D~\cite{eg3d} & 
\textbf{14.4} & \textbf{17.9} & \textbf{0.70} & \textbf{1.32} &  %
\textbf{15.0} & \textbf{20.9} & 0.61 & 1.02 &  %
\textbf{10.5} & \textbf{14.2} & 0.71 & 1.18 &  %
\textbf{13.3} & \textbf{17.7} & 0.67 & \textbf{1.17} &  %
\textbf{15.6} & \textbf{19.6} & 0.97 & \textbf{0.90} \\ %
Ours & 
42.1 & 46.5 & 0.46 & 0.26 &  %
38.5 & 43.5 & \textbf{0.84} & \textbf{1.31} &  %
48.0 & 53.3 & \textbf{0.85} & \textbf{1.46} &  %
42.8 & 47.8 & \textbf{0.72} & 1.01 &  %
15.7 & \textbf{19.6} & \textbf{0.99} & 0.65 \\ %
 \bottomrule   
\end{tabular}
}
\end{table*}

\begin{table}[t]
\centering
\caption{\small 3D generation performance for our model RenderDiffusion, and baselines GIRAFFE~\cite{niemeyer2021giraffe}, pi-GAN~\cite{chan2021pi} and EG3D~\cite{eg3d}, on FFHQ (faces) and AFHQ (cats). We report FID for train viewpoints, as well as coverage (\textit{cov.}) and density (\textit{dens.}) \cite{naeem2020reliable}. For GIRAFFE and pi-GAN FID, we use the results from \cite{eg3d}; for EG3D we use resolution $64\times64$, i.e.~the same as ours; we omit pi-GAN coverage and density due to lack of a publicly-available checkpoint on which to calculate these.}
\label{tab:generation-fhq}
\vspace{-8pt}
\scalebox{1}{%
\setlength{\tabcolsep}{2pt}%
\begin{tabular}{@{}lcccccc@{}}
\toprule  
& \multicolumn{3}{c}{FFHQ} & \multicolumn{3}{c}{AFHQ} \\
\cmidrule(lr){2-4} \cmidrule(l){5-7}
 & $\text{FID}_r$ & $\text{cov.}$ & $\text{dens.}$ & $\text{FID}_r$ & $\text{cov.}$ & $\text{dens.}$ \\
 \midrule
GIRAFFE~\cite{niemeyer2021giraffe} & 
31.5 & 0.66 & 1.17 & %
16.1 & 0.07 & 0.20 \\ %
pi-GAN~\cite{chan2021pi} & 
29.9 & -- & -- & %
16.0 & -- & -- \\ %
EG3D~\cite{eg3d} & 
19.8 & 0.68 & 1.20 & %
23.7 & 0.37 & 0.94 \\ %
Ours & 
59.3 & 0.31 & 1.01 & %
18.0 & 0.26 & 0.37 \\ %
 \bottomrule   
\end{tabular}
}
\end{table}

\section{Additional Unconditional Generation Results}
\label{sec:supp_unconditional}

\paragraph{Quantitative evaluation}

We evaluate the distributions of generated scenes quantitatively using four metrics. $\text{FID}_r$ is the Fr\'echet Inception Distance (FID)~\cite{heusel17fid} computed between training views of the generated scenes and training views of all training set scenes. $\text{FID}_t$ is the FID computed between test views of the generated scenes and test views of all scenes from the test set. 
The coverage metric (cov.)~\cite{naeem2020reliable} measures how well the generated distribution covers the data distribution. It is defined as the fraction 
of training set
images with neighborhoods that contain at least one generated sample, with a neighborhood defined based on the 3-nearest neighbors.
Similarly, the density metric (dens.)~\cite{naeem2020reliable} measures how close generated samples are to the data distribution, by calculating the average number of real samples whose neighborhoods contain each generated sample. Neighborhoods are defined in the feature space of a VGG-16 network (last hidden layer) that was applied to all training set views of a generated scene.
The latter two metrics are similar to the \textit{improved recall and precision} metrics of \cite{kynkaanniemi19ipr}, but avoid certain pathological behaviors~\cite{naeem2020reliable}.

Results on the synthetic datasets
are presented in \tab{generation}. We see that EG3D performs well on the FID metric, with ours second for CLEVR and pi-GAN second for ShapeNet. Our approach tends to perform better on the coverage and density metrics, while pi-GAN is particularly poor on these. To interpret our quantitative performance relative to EG3D, we refer to the qualitative results shown in Figure 5 of the main paper, where we see that our shapes are slightly more blurry than EG3D, but exhibit more variety and similar shape quality, apart from the blurriness. We hypothesize that the blurriness introduces a bias that the FID metric is highly sensitive to (as the blurriness may affect the feature average that the FID is based on). Coverage and density are less sensitive to the blurriness, as they don't rely on an average over all samples and instead provide a more detailed comparison of the sample distributions by working with sample neighborhoods. This interpretation of the quantitative results suggests that, leaving aside the blurriness, our method generates samples that better cover the data distribution, at a comparable sample quality. This is in line with current understanding of the differences between diffusion models like \name, and GANs like EG3D. Note that our models were not fully converged at the time of measuring these results, and we observed that the blurriness gradually decreases over the course of the training, making it likely that the blurriness can be reduced with additional training.
\tab{generation-fhq} shows quantitative results on the real datasets FFHQ (photos of human faces) and AFHQ (photos of cat faces); see also the qualitative results in the main paper.

\begin{figure}
  \centering
    \includegraphics[width=\linewidth]{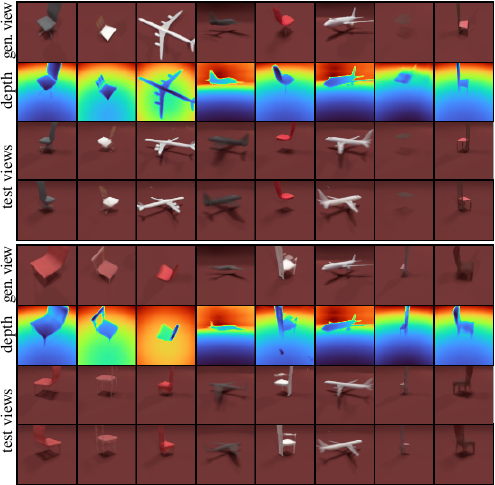}
  \caption{Multi-category generation results. We show generated scenes from a single \name model trained on both \texttt{chair} and \texttt{airplane} categories.}
  \label{fig:multicategory}
  \vspace{-1,5em}
\end{figure}

\begin{figure}
    \centering
    \includegraphics[width=0.48\linewidth]{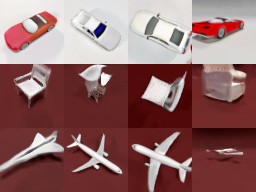}~
    \includegraphics[width=0.48\linewidth]{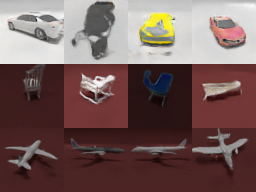}
    \caption{Uncurated samples from pi-GAN (left four columns) and GIRAFFE (right four columns), on the three ShapeNet classes. Compare with results from ours and EG3D in the main paper.}
    \label{fig:generation-pigan-giraffe}
\end{figure}

\paragraph{Additional qualitative results}
The supplementary video shows additional uncurated (not cherry-picked) qualitative results for unconditional generation, shown from a camera that rotates around the object.
\fig{generation-pigan-giraffe} shows uncurated generated samples from GIRAFFE and pi-GAN, on the three ShapeNet classes.

\begin{table*}
\centering
\caption{\small 3D reconstruction performance when part of the input image is masked. For easier comparison, we copy the unmasked results from the table in the main paper.}
\label{tab:supp_inpainting_table}
\scalebox{0.96}{
\begin{tabular}{@{}lcccccccccc@{}}
\toprule  
& \multicolumn{8}{c}{ShapeNet} & \multicolumn{2}{c}{CLEVR1} \\
\cmidrule(lr){2-9} \cmidrule(l){10-11}
& \multicolumn{2}{c}{\texttt{car}} & \multicolumn{2}{c}{\texttt{plane}} & \multicolumn{2}{c}{\texttt{chair}} & \multicolumn{2}{c}{average} \\
\cmidrule(lr){2-3} \cmidrule(lr){4-5} \cmidrule(lr){6-7} \cmidrule(lr){8-9}
 & PSNR & SSIM & PSNR & SSIM & PSNR & SSIM & PSNR & SSIM & PSNR & SSIM\\
 unmasked input & \textbf{25.4} & \textbf{0.805}  & 26.3 & 0.834  & \textbf{26.6} & \textbf{0.830}  & 26.1 & 0.823  & \textbf{39.8} & \textbf{0.976}  \\
 masked input & 24.7 & 0.790  & \textbf{27.6} & \textbf{0.870}  & 26.2 & 0.820  & \textbf{26.2} & \textbf{0.827}  & 38.9 & 0.970  \\
 \bottomrule   
\end{tabular}
}
\end{table*}

\section{Additional Inpainting Results}
\label{sec:supp_inpainting}

To quantitatively measure how well our generative model inpaints 3D scenes, we treat inpainting as a 3D reconstruction task with occlusions, where the mask is the occluder. Similar to the unoccluded case, we compare renders of the reconstructed scene from test set viewpoints to ground truth renders using PSNR and SSIM as metrics.
Since there are often many plausible inpaintings (i.e.~the task is ambiguous), we sample $K$ different inpaintings with our model and select the best matching one. For CLEVR we choose $K=25$ as the mask often hides majority of the object (increasing the degree of ambiguity), while for ShapeNet we choose $K=10$. This gives as an indication if the distribution of generated scenes for a given masked input image includes the ground truth scene.
To choose the masked-out region of each image, we use a square with width and height equal to 40\% of the image resolution (e.g. for an image of size $64\times64$ the mask will be of size $26\times26$). The mask is placed uniformly at random within a square region of side length 
$\frac{5}{16}$
of the image size, itself centered in the image.
This ensures the mask always covers part of the foreground object, not just the background. Illustration of masked inputs and diversity in \name predictions is shown in \fig{inpainting}.

Quantitative results on this task are given in \tab{supp_inpainting_table}. We compare the reconstruction performance with and without masked input. We can see that in most cases, the performance for the two cases is comparable, indicating that a scene resembling the ground truth is contained in the output distribution.
We show additional qualitative results with multiple seeds in the supplementary video.

\begin{figure}[t]
  \centering
    \includegraphics[width=\linewidth]{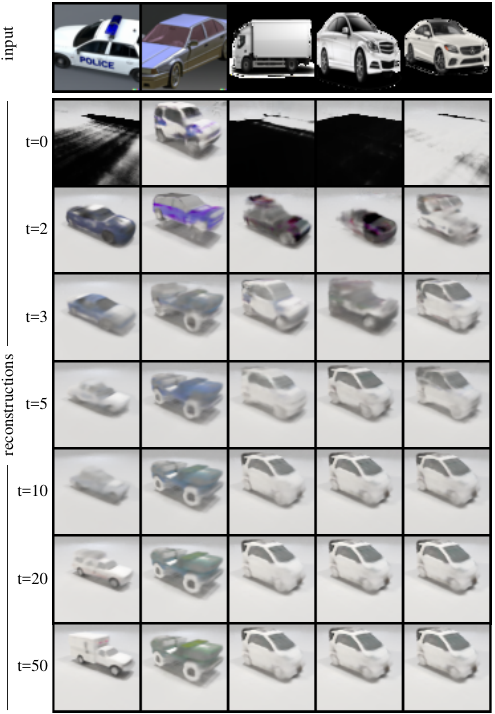}
  \caption{Additional reconstructions from out-of-distribution images. We show reconstructions with different amounts of added noise for out-of-distribution input. We use the same random seed for all reconstructions. Note how the amount of noise trades of between reconstruction quality and fidelity to the input image.}
  \label{fig:supp_oodinference}
  \vspace{-1,5em}
\end{figure}

\section{Multi-Category Generation Results}
\label{sec:supp_multimodal}

To further demonstrate that our model can represent complex, multi-modal distributions, we perform an additional experiment where a single model is trained jointly on multiple ShapeNet categories.
Specifically, we train \name on the union of the \texttt{chair} and \texttt{plane} categories, otherwise using the same architecture and training protocol as described in the main paper.

In \fig{multicategory}, we show qualitative results from this model.
We see that \name has successfully captured both modes of the dataset, sampling plausible chairs and airplanes.
As in the single-category experiments in the main paper, the samples are 3D-consistent, exhibit plausible depth-maps, and look realistic from novel test viewpoints.

\section{Additional Reconstruction Results}
\label{sec:supp_oodinference}

\begin{figure*}[t]
  \centering
\includegraphics[width=\linewidth]{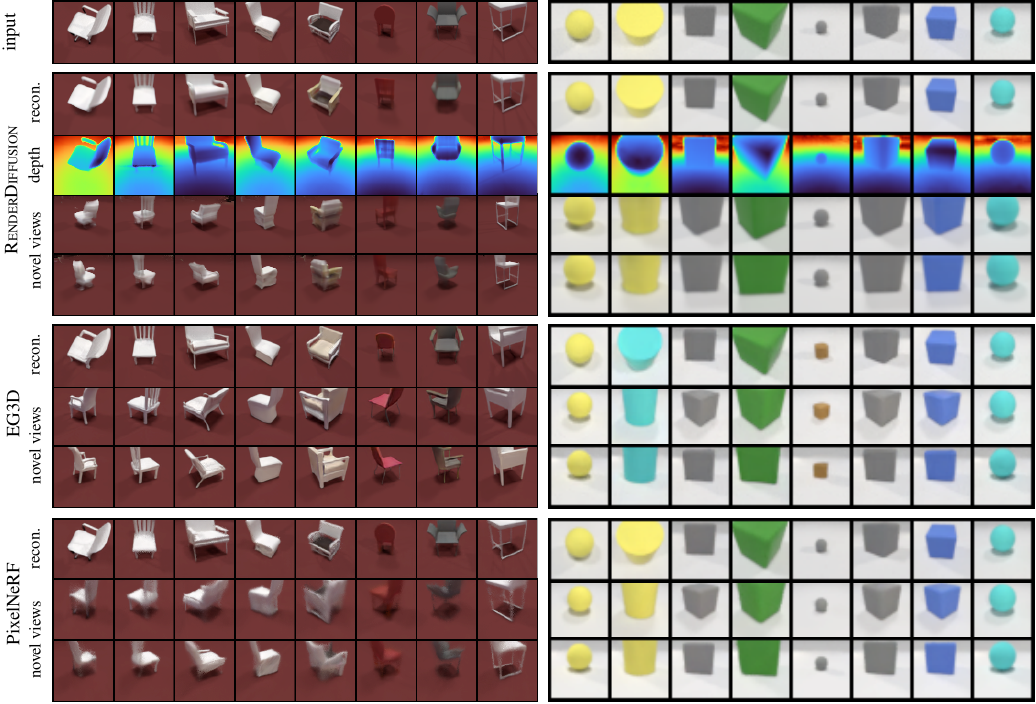}
  \caption{Reconstruction results for ShapeNet \texttt{chair} and CLEVR1. Similar to the results on \texttt{car} and \texttt{plane} datasets in the main paper, our reconstructions better preserve shape identity than EG3D, and are sharper and more detailed than PixelNeRF.}
  \label{fig:reconstruction2}
  \vspace{-1em}
\end{figure*}

In Figure~\ref{fig:reconstruction2}, we show addditional reconstruction results for CLEVR1 and ShapeNet \texttt{chair} datasets. 
In Figure~\ref{fig:supp_oodinference}, we show reconstruction from out-of-distribution images with different amounts of added noise, ranging from no noise at $t=0$ to $50$ noise steps at $t=50$. Adding larger amounts of noise results in reconstructions that are more generic and increasingly diverge from the input image, as the generative model fills in details covered by the noise, but also show increasingly higher-quality shapes.

\end{document}